\documentclass[conference]{IEEEtran}
\IEEEoverridecommandlockouts
\usepackage{cite}
\usepackage{amsmath,amssymb,amsfonts}
\usepackage{algorithmic}
\usepackage{graphicx}
\usepackage{textcomp}
\usepackage{xcolor}
\def\BibTeX{{\rm B\kern-.05em{\sc i\kern-.025em b}\kern-.08em
    T\kern-.1667em\lower.7ex\hbox{E}\kern-.125emX}}
\begin{document}

\title{On the Universality of Transformer Architectures; How Much Attention Is Enough?\\
}

\author{
\IEEEauthorblockN{Amirreza Abbasi}
\IEEEauthorblockA{\textit{Department of Computer Science and Information Technology}\\
\textit{Institute for Advanced Studies in Basic Sciences (IASBS)}\\
Zanjan, Iran\\
a.abbasi@iasbs.ac.ir}
\and
\IEEEauthorblockN{Mohsen Hooshmand}
\IEEEauthorblockA{\textit{Department of Computer Science and Information Technology}\\
\textit{Institute for Advanced Studies in Basic Sciences (IASBS)}\\
Zanjan, Iran\\
mohsen.hooshmand@iasbs.ac.ir}
}

\maketitle

\begin{abstract}
Transformers are crucial across many AI fields, such as large language models, computer vision, and reinforcement learning. This prominence stems from the architecture’s perceived universality and scalability compared to alternatives. This work examines the problem of universality in Transformers, reviews recent progress, including architectural refinements such as structural minimality and approximation rates, and surveys state-of-the-art advances that inform both theoretical and practical understanding. Our aim is to clarify what is currently known about Transformers expressiveness, separate robust guarantees from fragile ones, and identify key directions for future theoretical research.
\end{abstract}


\section{Introduction}
Transformers have established their revolutionary role in deep learning through their core power: attention mechanisms \cite{b1}, which help models learn via parameter sharing and parallel token processing. This architectural foundation also underlies modern large language models (LLMs), whose remarkable empirical capabilities further amplify the importance of understanding transformer expressiveness at a theoretical level. Expressiveness refers to the inherent capacity of a model class to represent a wide variety of functions or mappings \cite{b2}. In the context of Transformers, it captures the range of sequence-to-sequence behaviors the architecture can realize under given architectural constraints, independent of training or parameter efficiency \cite{b2}\cite{b3}.  Researchers analyze the expressivity of Transformers from two complementary aspects: universal function approximation, which asks whether Transformers can approximate arbitrary continuous sequence-to-sequence mappings on compact domains to arbitrary precision. Another perspective comes from formal language theory and complexity classes, which study what Transformers can compute under explicit resource constraints \cite{b3}.

This survey primarily adopts the approximation-theoretic lens, while drawing connections to complexity based limitations when they reveal fundamental expressiveness gaps. Early universality results established that standard Transformers are universal approximators of permutation equivariant sequence-to-sequence functions \cite{b2}, and this property extended to other architectures modified in terms of efficiency and size, with expressiveness preserved.

Alongside these achievements, Transformers exhibited fundamental limitations in approximating certain functions and general expressiveness \cite{b10}, which have been carefully analyzed. Finally, recent advancements in prompting and parameter-efficient tuning have been studied \cite{b12}, providing insights into the expressiveness of these Transformers variants.

As a contribution, this survey provides a unified and structured view of expressiveness results for Transformers architectures under the approximation lens. Specifically, we synthesize classical universality results and highlight the core proof techniques that support later developments. We also demonstrate Transformers universality robustness across various architectures, offering a road map for researchers in applied fields to select appropriate, efficient, and expressive Transformers.

By organizing these findings within a common framework, this survey clarifies what is known about Transformers expressiveness, highlights open challenges, and points toward promising directions for future theoretical research. 

The remainder of the paper is organized as follows. Section ~\ref{sec:pre} introduces necessary notation and preliminary definitions. Section ~\ref{sec:main} reviews foundational universality results and efficient attention mechanisms. Section ~\ref{sec:beyond} discusses extensions beyond classical universality, including minimal architectures, approximation rates, and new input regimes. Section~\ref{sec:prompt} focuses on parameter-efficient universality through prompting followed by open theoretical questions and conclusion in ~\ref{sec: con}.

\section{Preliminaries and definitions} \label{sec:pre}
In this section, we introduce key concepts, notation, and definitions used throughout this survey. In order to be intuitive and open the doors to this side of theoretical computing of Transformers, we have refrained from explaining the proofs in more detail, especially in their mathematical form. So the prerequisites are presented merely to further clarify and unify the common understanding of the concepts.\\
A function $f: \mathbb{R}^{d\times n} \to \mathbb{R}^{d\times n}$ is permutation equivariant if, for any permutation $\pi$ of $[n]$, we have \[
f(\pi \cdot x) = \pi \cdot f(x), \text{where } \pi \cdot x = (x_{\pi(1)}, \dots, x_{\pi(n)}).
\] Permutation equivariant is an important inherent property of Transformers without positional encoders. This ensures the output depends only on the set of input tokens, not their order, and it is key for proving universality of Transformers without positional encoders.
Throughout this survey, we use sequence-to-sequence expression a lot, which formally can interpret as function $f: \mathbb{R}^{d\times n}\to \mathbb{R}^{d\times n}$. 
A Transformers $\mathcal{T}$ is a universal approximator of a class of continuous functions $\mathcal{F}$ if, for any $f \in \mathcal{F}$ and any $\epsilon > 0$, there exist Transformers parameters such that

\[
\| \mathcal{T}(x) - f(x) \| < \epsilon, \quad \forall x \in \mathcal{X},
\]

\noindent where $\mathcal{X}$ is a compact input domain. This property is central to evaluating Transformers expressiveness under architectural modifications.

\subsection{Transformers architecture}
A (vanilla) Transformers is a sequence-to-sequence model composed of stacked layers, each consisting of a multi head self-attention mechanism followed by a token-wise feed-forward network (FFN). Given an input sequence $X = (x_1, \dots, x_n)$, the model first applies an embedding map and for certain Transformers, a positional encoding.

In each self-attention layer, queries, keys, and values are computed via linear projections $Q = X W^Q$, $K = X W^K$, and $V = X W^V$, and attention outputs are formed using a softmax weighted aggregation of values, implementing contextual mapping, where each token’s representation is updated based on information aggregated from all other tokens. Multi head attention concatenates multiple such mechanisms with independent parameters. The FFN is applied identically and independently to each token.

A Transformers is parameterized by its depth (number of layers), width (hidden dimension), number of attention heads, and sequence length. Variants modify attention structure, parameterization, or aggregation while preserving this overall framework.

\section{Expressiveness} \label{sec:main}
This section synthesizes existing results to show that the universality of Transformers is robust to architectural constraints. Additionally,  we explain efficiency mechanisms such as sparsity, low-rank projections, and kernelized attention preserve expressiveness as long as global information routing is maintained.

\begin{table*} 
\centering
\caption{Efficiency--Expressiveness Trade-offs in Transformers Architectures}
\label{tab:efficiency_expressiveness}
\renewcommand{\arraystretch}{1.15}
\begin{tabular}{l c c c c c c}
\hline
\textbf{Architecture} 
& \textbf{Attention Pattern} 
& \textbf{Depth} 
& \textbf{Heads} 
& \textbf{Parameter Reduction} 
& \textbf{Universality} 
& \textbf{Approximation Rate} \\
\hline
Vanilla Transformers 
& Dense 
& Deep 
& Multi 
& -- 
& Yes 
& Not explicit \\

Sparse Transformers 
& Sparse 
& Moderate 
& Multi 
& $O(n)$ connections 
& Yes 
& Not explicit \\

Low-rank Transformers 
& Dense 
& 1 
& 1 
& Strong 
& Yes 
& Implicit \\

Softmax-only Transformers 
& Dense 
& 2 
& Multi 
& Moderate 
& Yes 
& Improves with heads \\

Performer 
& Kernelized 
& Shallow 
& Multi 
& Strong 
& Yes 
& Implicit \\

Sumformer 
& Global summation 
& 1 
& -- 
& Strong 
& Yes 
& Latent-dimension trade-off \\

Prefix-tuned Transformers 
& Frozen + prefix 
& Linear 
& 1 
& Very strong 
& Yes 
& Not studied \\
\hline
\end{tabular}
\end{table*}

\subsection{Expressivity of vanilla Transformers}
First fundamental results regarding the expressiveness of Transformers were proposed by Yun et al. \cite{b2}. They proved that Transformers are universal approximators for continuous permutation equivariant sequence-to-sequence functions on compact domains, and then they extended this property to the Transformers with positional encoders. The core idea of the proof proceeds through three consecutive approximation steps: first approximating a continuous target function by a piecewise constant function, then representing this piecewise constant function via intermediate token-wise constructions, and finally realizing the resulting mapping using a Transformers architecture. In more intuitive way, continuous function are quantize into small fixed cubes that facilitate approximation for Transformers. The universality result is nontrivial given architectural constraints such as shared parameters and token-wise feed-forward processing. But the key to overcoming this issue is formalizing and showing the power of attention in contextual mapping. Despite the token-wise nature of the feed-forward layers, with sufficient depth, attention enables global information flow by tagging and shifting token representation to simulate any piecewise constant function. This suggests that, within the theoretical setting, observed expressiveness limitations are primarily architectural rather than fundamental, which arise from practical constraints such as finite width, depth, and trainability. These considerations motivated subsequent work on efficient Transformers variants, whose expressiveness is analyzed in later sections based on the same proof procedure.

\subsection{Expressivity of sparse Transformers}
Quadratic computation complexity bottlenecks of vanilla Transformers have been addressed by providing an attention graph, treating each token as a node in graph and limits the connecting edges of tokens to reduce complexity \cite{b17}. Building on Yun et al.’s universality framework for sequence-to-sequence functions, subsequent work established the same results for sparse Transformers \cite{b3}. The proof again relies on the sequence of approximations method introduced by Yun et al. But since there are several sparse attention patterns with structurally different connectivity graphs, a unifying framework was essential to characterize expressiveness independently of any specific design. This is achieved via a minor modification of the attention layer that abstracts attention as a graph based routing mechanism, from which three sufficient conditions on any sparse pattern are derived to guarantee universality. These conditions include: self information preservation, i.e., each token always attends to itself, global connectivity that can be translated as the existence of an ordered Hamiltonian path among tokens, and global information propagation guarantees the existence of direct or indirect connections between all tokens after a sparse attention layer.
As some sparse patterns such as Star \cite{b13} and BigBird \cite{b14} only need O(n) connections to satisfy conditions, it turns out that “O(n) connections are expressive enough” \cite{b3}. A significant result shows that, using a fixed number of additional layers, sparse attention can simulate a single dense attention layer. This simulation relies on routing rather than all to all connectivity within a single layer. These theoretical insights are further supported by empirical evaluations reported in the same work.

\subsection{Expressivity of efficient Transformers; Sumformer}
Beyond sparsity, efficient variants like Linformer \cite{b15} (low-rank factorization) and Performer \cite{b16} (kernel based softmax approximation) reduce complexity to O(nk) and O(nkd), respectively. Sils et al. \cite{b4} introduced a Sumformer framework which is a tool to prove universal approximation not only for these efficient Transformers, but for vanilla Transformers as well. The core idea of Sumformer is using algebraic properties \cite{b18}. Multisymmetric polynomials are permutation invariant functions, and simulating them by the Sumformer mechanism enables universal approximation.
Sumformer replaces pairwise attention with a global summation over transformed inputs. This allows each token to condition on sequence level statistics while preserving permutation structure and is used as a tool to show other architectures universality by simulating Sumformer aggregation under suitable parameterizations. Since it sums over transformed input and conditions each token based on this sum, invariance for global aggregation is satisfied. Equivariance for token-wise outputs is also satisfied, which is essential for universal approximation. The universality result comes with a fundamental trade off between representational resources. While approximation of continuous functions necessitates a large latent dimension, discrete function approximation can be achieved with a one dimensional latent space, provided the complexity is shifted to the feed-forward layers through exponential growth.
The contribution of the paper is showing universal approximation of Performer and Linformer along with new a notable result of vanilla Transformers shown to be a universal approximator with just one layer of self-attention. 

\section{Beyond classical universality results} \label{sec:beyond}

Results of ~\ref{sec:main} establish that Transformers remain universal under modified architectural assumptions. Although they leave open questions regarding efficiency, quantitative rates, and practical constraints which this section discusses them.

\subsection {Architectural constraints and minimality}
Due to computational and parameter efficiency constraints in practice, minimality of models is a major concern in the field. Minimality refers to the property that a model achieves a target expressive or approximation capability such as universal approximation using the smallest possible architectural resources, measured in terms of depth, width, number of attention heads \cite{b19}. Yun’s result is existential and resource heavy, hence distant from practical architectures. In \cite{b5}, the authors showed that universality can be achieved with a single attention layer with one head, followed by FFN. In \cite{b5} while the permutation equivariance and universal approximation property is preserved, low-rank weight matrices used instead of full rank $Q,K,V,O$ matrices. This method effectively projecting the high dimensional token representations into a smaller subspace, which reduces the number of parameters while maintaining sufficient expressive capacity.\\ 
A stronger minimality result appeared in \cite{b6}, which made one of the most fundamental modifications to Transformers. They showed that a two layer self-attention only architecture can serve as a universal approximator. In this modification, interpolation relies on using attention weights, consequently FFNs are not strictly necessary. By adding more attention heads, one can improve approximation.  
In another significant result, authors enriched their work by stating that adding softmax can simulate interpolation and enable universality with a single attention layer under specific constructions. These results highlighted the central role of attention in Transformers, which can approximate piecewise linear functions and simulate interpolation methods. 

\subsection{Efficient without loss of expressiveness}
Prior works established that Transformers are universal approximators (expressivity). Quantifying how efficiently they can approximate sequence-to-sequence functions, Jiang and li~\cite{b7} are pioneers in providing explicit approximation rate bounds for Transformers by linking the error decay to model capacity and structural properties of target sequences. 
The core idea is based on Jackson type approximation theorem~\cite{b21}, showed how the approximation error decreases as feed-forward width, attention dimension, and sequence length grow. Higher approximation rates allow achieving a target error with fewer model parameters. Understanding dependencies of approximation rate helps to design more efficient Transformers. The authors introduced precise bounds for Transformers, enabling comparisons with prior architectures like RNNs and guiding design choices for parameter efficient models. Additionally, they offer a formal comparison with RNNs, highlighting that Transformers are inherently more parameter efficient at capturing long range dependencies. This result suggests that the Transformers' success is not merely due to its capacity, but its ability to reach target accuracies with lower architectural complexity than recurrent alternatives. 
These promising approximation results comparing to other deep networks can contribute to a deeper theoretical understanding of Transformers and their ability. 

\subsection{New input regimes and dimensionality}
A primary challenge in sequence modeling is the curse of dimensionality, where the complexity of approximation can grow exponentially with input dimension. However, these results indicate that high dimensional inputs do not preclude the universality of Transformers. Prior works suggest that feed-forward layers can overcome this issue, and in \cite{b8} the authors showed that representing Transformers via FFNs preserves the same universality guarantees.
While previous studies had defined $\ell_{p}, \text{where } p<\infty$, this work has another notable contribution of introducing strict $\ell_{\infty}$ error bounds for sequence-to-sequence mappings, demonstrating that Transformers achieve uniform approximation across the entire input domain, rather than only minimizing average error. In this setting, the role of attention differs from that in vanilla Transformers, it functions as an operator implementing functional transformations over the sequence, rather than primarily performing semantic contextualization which is updates token representations to reflect meaning and relationships within a sequence.

A complementary study, \cite{b9}, further demonstrated that by leveraging feature extraction and parameter sharing properties, Transformers can effectively escape the curse of dimensionality. Together, these works confirmed that Transformers architectures maintain strong expressiveness even in very high dimensional sequence spaces.

\subsection{Limits of expressiveness} 
Not all expressiveness results are favorable to Transformers. While many architectures enjoy universal approximation guarantees, certain design choices provably limit expressiveness. In particular, Luo et al. \cite{b10} showed that Transformers equipped with relative positional encoders (RPEs) \cite{b20} fail to approximate certain continuous sequence-to-sequence functions under standard architectural constraints. This is quite surprising since RPE is efficient version of absolute positional encoding which uses relative distance between pairs of tokens and the initial assumption is on its expressiveness. The failure arises from the placement of RPE terms inside the softmax attention mechanism, where positional biases interact multiplicatively with content based attention weights and cause translation invariance. Thus, the model cannot distinguish absolute token positions, which is required for many non invariant tasks. The same work further proposes a modified architecture that restores universality.

Nath et al.~\cite{b11} found  another limitations in regression settings: although Transformers are universal approximators, they exhibit poor expressiveness when approximating smooth functions. Even full encoder$/$decoder architectures exhibit poor approximation rates in worst case smooth regression tasks, requiring excessively large model sizes. These results highlight a gap between theoretical universality and practical approximation efficiency.

\section{Prompting and parameter-efficient universality}
\label{sec:prompt}
Recently, prompting has found its place in research on Transformers. Regarding expressiveness, the key question is: ``Can a pre-trained model, through prompt or prefix tuning, universally approximate a continuous sequence-to-sequence function?'' The answer is affirmative as shown in \cite{b12}, revises prior assumptions about pre-trained models showing that even smaller properly prefix tuned models can serve as universal approximators.
Prefixes are vectors that prepend to already frozen weight matrices in the self-attention layers. These prefix vectors modify the Transformers's effective parameters, enabling the simulation of a wide range of functions. In the universal approximation setting, prefix vectors act as controllers of the network’s parameters.
Regarding minimality, prefix Transformers require just one attention head to approximate any continuous function. The work further demonstrates that the required network depth scales linearly with sequence length, highlighting an efficient relationship between architecture and input size.
Although this paper demonstrates the theoretical power of prompting for universal approximation, several open questions remain. Importantly, the work provides a constructive proof but does not include optimization or empirical validation. Approximation guarantees of this work do not imply practical learnability, and the scalability of those prefix models remains unclear. Moreover, the effectiveness of prompting for discrete function approximation has not yet been explored, representing a key direction for future research.

\section{CONCLUSION AND OPEN QUESTIONS}\label{sec: con}
This survey has examined the expressiveness of Transformers architectures through the lens of approximation theory, synthesizing a growing body of results on universality, architectural efficiency, and fundamental limitations. By organizing classical universality theorems alongside more recent developments, including sparse attention, low-rank and kernelized variants, minimal architectures, approximation rates, high dimensional input regimes, and prompting based universality, we have shown that Transformers expressiveness is remarkably robust to architectural constraints, provided that global information routing is preserved.

From an applied perspective, these findings offer concrete guidance for model design. Theoretical guarantees indicate that architectural choices such as sparsity patterns, low-rank projections, reduced attention heads, or parameter efficient tuning can substantially lower computational and parameter costs without sacrificing expressiveness. As a result, practitioners can leverage these insights as shown in table~\ref{tab:efficiency_expressiveness} to select Transformers variants that are better aligned with practical constraints on memory, computation, and data availability, while retaining strong approximation capabilities.

At the same time, the surveyed results expose several open theoretical challenges. Universality guarantees are often existential and silent on optimization, leaving open questions about learnability, training dynamics, and sample efficiency. Quantitative approximation rates remain incomplete, particularly beyond smooth function classes or under realistic architectural restrictions. Moreover, expressiveness under discrete settings, strict resource bounds, or alternative positional encoding schemes is not yet fully understood. These gaps indicate that, despite significant progress, the theoretical foundations of Transformers architectures are still evolving.

By connecting approximation theoretic insights with architectural design choices, we hope this work serves both as a reference for applied researchers and as a roadmap for future investigations into the theoretical limits of attention-based models.

\vspace{12pt}


\begin{thebibliography}{00}
\bibitem{b1} Vaswani, Ashish, Noam Shazeer, Niki Parmar, Jakob Uszkoreit, Llion Jones, Aidan N. Gomez, Łukasz Kaiser, and Illia Polosukhin. "Attention is all you need." Advances in neural information processing systems 30 (2017).
\bibitem{b2} Yun, Chulhee, Srinadh Bhojanapalli, Ankit Singh Rawat, Sashank Reddi, and Sanjiv Kumar. "Are Transformers universal approximators of sequence-to-sequence functions?." In International Conference on Learning Representations.
\bibitem{b3} Yun, Chulhee, Yin-Wen Chang, Srinadh Bhojanapalli, Ankit Singh Rawat, Sashank Reddi, and Sanjiv Kumar. "O (n) connections are expressive enough: Universal approximability of sparse Transformers." Advances in Neural Information Processing Systems 33 (2020): 13783-13794.
\bibitem{b4} Alberti, Silas, Niclas Dern, Laura Thesing, and Gitta Kutyniok. "Sumformer: Universal approximation for efficient Transformers." In Topological, Algebraic and Geometric Learning Workshops 2023, pp. 72-86. PMLR, 2023.
\bibitem{b5} Kajitsuka, Tokio, and Issei Sato. "Are Transformers with One Layer self-attention Using Low-Rank Weight Matrices Universal Approximators?." In The Twelfth International Conference on Learning Representations.
\bibitem{b6} YHu, Jerry Yao-Chieh, Hude Liu, Hong-Yu Chen, Weimin Wu, and Han Liu. "Universal Approximation with Softmax Attention." arXiv preprint arXiv:2504.15956 (2025).
\bibitem{b7} Jiang, Haotian, and Qianxiao Li. "Approximation rate of the Transformers architecture for sequence modeling." Advances in Neural Information Processing Systems 37 (2024): 68926-68955.
\bibitem{b8} Jiao, Yuling, Yanming Lai, Yang Wang, and Bokai Yan. "Transformers Can Overcome the Curse of Dimensionality: A Theoretical Study from an Approximation Perspective." arXiv preprint arXiv:2504.13558 (2025).
\bibitem{b9} Takakura, Shokichi, and Taiji Suzuki. "Approximation and estimation ability of Transformers for sequence-to-sequence functions with infinite dimensional input." In International Conference on Machine Learning, pp. 33416-33447. PMLR, 2023.
\bibitem{b10} Luo, Shengjie, Shanda Li, Shuxin Zheng, Tie-Yan Liu, Liwei Wang, and Di He. "Your Transformers may not be as powerful as you expect." Advances in Neural Information Processing Systems 35 (2022): 4301-4315.
\bibitem{b11} Nath, Swaroop, Harshad Khadilkar, and Pushpak Bhattacharyya. "Transformers are Expressive, But Are They Expressive Enough for Regression?." arXiv preprint arXiv:2402.15478 (2024).
\bibitem{b12} Petrov, Aleksandar, Philip Torr, and Adel Bibi. "Prompting a Pretrained Transformers Can Be a Universal Approximator." In Forty-first International Conference on Machine Learning.
\bibitem{b13} Guo, Qipeng, Xipeng Qiu, Pengfei Liu, Yunfan Shao, Xiangyang Xue, and Zheng Zhang. "Star-Transformers." In Proceedings of the 2019 Conference of the North American Chapter of the Association for Computational Linguistics: Human Language Technologies, Volume 1 (Long and Short Papers), pp. 1315-1325. 2019.
\bibitem{b14} Zaheer, Manzil, Guru Guruganesh, Kumar Avinava Dubey, Joshua Ainslie, Chris Alberti, Santiago Ontanon, Philip Pham et al. "Big bird: Transformers for longer sequences." Advances in neural information processing systems 33 (2020): 17283-17297.
\bibitem{b15} Wang, Sinong, Belinda Z. Li, Madian Khabsa, Han Fang, and Hao Ma. "Linformer: Self-attention with linear complexity." arXiv preprint arXiv:2006.04768 (2020).
\bibitem{b16} Choromanski, Krzysztof Marcin, Valerii Likhosherstov, David Dohan, Xingyou Song, Andreea Gane, Tamas Sarlos, Peter Hawkins et al. "Rethinking Attention with Performers." In International Conference on Learning Representations.
\bibitem{b17} Zucchet, Nicolas, Francesco d'Angelo, Andrew K. Lampinen, and Stephanie CY Chan. "The emergence of sparse attention: impact of data distribution and benefits of repetition." arXiv preprint arXiv:2505.17863 (2025).
\bibitem{b18} Hutter, Marcus. "On representing (anti) symmetric functions." arXiv preprint arXiv:2007.15298 (2020).
\bibitem{b19} Bermeitinger, Bernhard, Tomas Hrycej, Massimo Pavone, Julianus Kath, and Siegfried Handschuh. "Reducing the Transformers Architecture to a Minimum." arXiv preprint arXiv:2410.13732 (2024).
\bibitem{b20} Shaw, Peter, Jakob Uszkoreit, and Ashish Vaswani. "Self-attention with relative position representations." arXiv preprint arXiv:1803.02155 (2018).
\bibitem{b21} Jackson, Dunham. The theory of approximation. Vol. 11. American Mathematical Soc., 1930.
\end{thebibliography}
\end{document}